%% file: root_arxiv.tex
\documentclass[letterpaper, 10 pt, conference]{ieeeconf}

\IEEEoverridecommandlockouts                              

\usepackage[utf8]{inputenc}

\usepackage{graphicx}
\graphicspath{/figs}
\usepackage{xcolor}
\usepackage{svg}
\usepackage{multirow}
\usepackage{booktabs}
\usepackage[algoruled,vlined,linesnumbered]{algorithm2e}
\usepackage{eso-pic}

\usepackage{amsmath}
\usepackage{amssymb}
\usepackage{subcaption}
\usepackage{todonotes}
\usepackage{mathtools}

\makeatletter
\let\NAT@parse\undefined
\makeatother
\usepackage[pdfa,colorlinks,bookmarksopen,bookmarksnumbered,allcolors=blue]{hyperref}
\usepackage{cite}

\usepackage{tikz}
\usetikzlibrary{automata,positioning,angles,quotes,calc}

\usepackage[
activate   = {true},
protrusion = true,
expansion  = true,
kerning    = true,
spacing    = true,
tracking   = false,
auto       = true,
selected   = true,
factor     = 1000,
stretch    = 20,
shrink     = 20,
]{microtype}

\newcommand{\ltl}{\textsc{ltl}\xspace}
\newcommand{\ltlf}{$\textsc{ltl}_f$\xspace}

\newcommand{\ltlfphi}{\phi}

\usepackage{amsthm}

\newtheoremstyle{main}
                {1em}                                                
                {1em}                                              
                {\normalfont}                                        
                {0pt}                                                
                {\scshape}                                           
                {\\*}                                                
                {2pt}                                                
                {\thmname{#1}\thmnumber{ #2}: \thmnote{\itshape #3}} 

\newtheorem{definition}{Definition}

\newtheorem{problem}{Problem}
\newtheorem{example}{Example}
\newtheorem{remark}{Remark}

\newcommand{\placetextbox}[3]{
  \setbox0=\hbox{#3}
  \AddToShipoutPictureFG*{
    \put(\LenToUnit{#1\paperwidth},\LenToUnit{#2\paperheight}){\vtop{{\null}\makebox[0pt][c]{#3}}}%
  }%
}%

\newif\ificra
\icratrue  
\icrafalse
\ificra
\fi

\newif\ifarxiv
\arxivtrue  
\ifarxiv
\fi

\title{\LARGE \bf Stochastic Games for Interactive Manipulation Domains}

\author{Karan Muvvala$^{*1}$, Andrew M. Wells$^{*2}$, Morteza Lahijanian$^1$, Lydia E. Kavraki$^2$, and Moshe Y. Vardi$^2$ 
\thanks{This work was supported in part by NASA 80NSSC21K1031, NASA 80NSSC17K0162, NSF 1830549, and NSF RI 2008720.
Authors would also like to thank the authors of PRISM, especially Dr. Dave Parker for their excellent tool and for their help in modifying it to import stochastic games. }
\thanks{$^*$Equal contribution}
\thanks{$^1$Aerospace Eng. Sciences Dept. at the University of Colorado Boulder. \texttt{firstname.lastname@colorado.edu}}
\thanks{$^2$Computer Science Dept. at Rice University. Andrew Wells was a student at Rice University at the time this work was conducted.
\texttt{andrewmw94@gmail.com}, 
\texttt{\{kavraki,vardi\}@cs.rice.edu} }
}

\begin{document}

\ifarxiv
\placetextbox{0.5}{0.95}{To appear at the 2024 IEEE International Conference on Robotics and Automation (ICRA), May 2024.}
\fi

\maketitle
\thispagestyle{plain}
\pagestyle{plain}

\input{sections/abstract}
\input{sections/introduction}
\input{sections/problem_setup}
\input{sections/modeling}
\input{sections/synthesis}
\input{sections/results}
\input{sections/conclusion}

\bibliographystyle{IEEEtran}
\bibliography{IEEEabrv,refs}

\end{document}

%% file: sections/abstract.tex
\begin{abstract}
As robots become more prevalent, the complexity of robot-robot, robot-human, and robot-environment interactions increases. In these interactions, a robot needs to consider not only the effects of its own actions, but also the effects of other agents' actions and the possible interactions between agents. Previous works have considered reactive synthesis, where the human/environment is modeled as a deterministic, adversarial agent; as well as probabilistic synthesis, where the human/environment is modeled via a Markov chain. While they provide strong theoretical frameworks, there are still many aspects of human-robot interaction that cannot be fully expressed and many assumptions that must be made in each model. In this work, we propose \emph{stochastic games} as a general model for human-robot interaction, which subsumes the expressivity of all previous representations. In addition, it allows us to make fewer modeling assumptions and leads to more natural and powerful models of interaction. We introduce the semantics of this abstraction and show how existing tools can be utilized to synthesize strategies to achieve complex tasks with guarantees.
Further, we discuss the current computational limitations and improve the scalability by two orders of magnitude by a new way of constructing models for PRISM-games.
\end{abstract}

%% file: sections/introduction.tex
\section{Introduction}

Traditionally, robots have accomplished complex tasks through \emph{planning} i.e., computing a ``path'' from the initial state to a goal state. As robots become more prevalent, the complexity of robot-robot, robot-human, or robot-environment interactions increases. In these interactions, a robot needs to consider not only the effects of its own actions, but also the effects of other agents' actions as well as the possible interactions between agents. These complexities mean \emph{planning} is often insufficient. Instead the robots should compute a \emph{strategy}, anticipating the possible effects of each agent's actions and reasoning in advance how it should respond to different contingencies.

\begin{figure}[t]
  \centering
  \includegraphics[width=0.6\columnwidth]{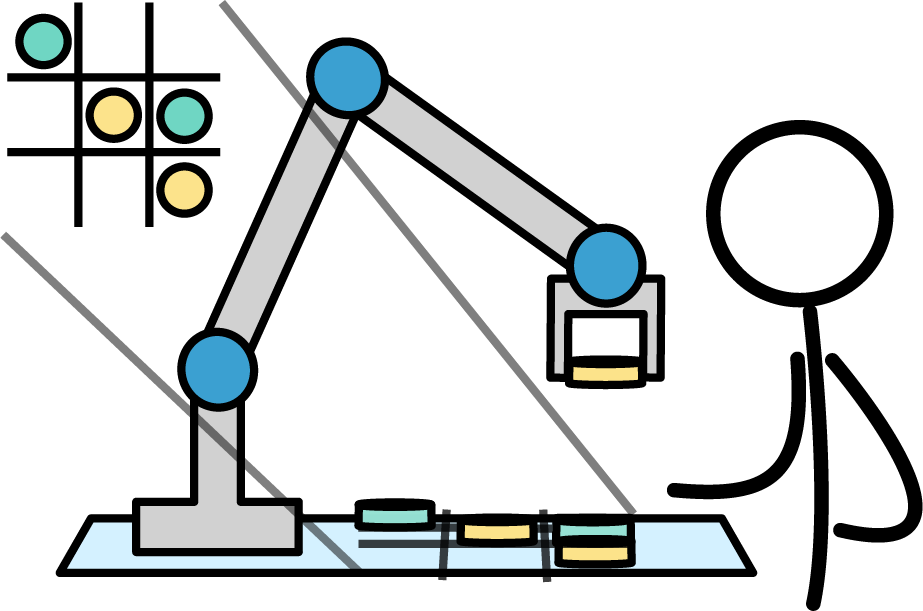}
    \caption{
    Tic-tac-toe game between a robot and a human.  The robot is unaware of the level of expertise of the human and suffers from the ``trembling hand'' problem. 
    In this case, the robot needs to reason about the probabilities of reaching a given state as well as the strategic responses of both agents from that state. 
    }
  \label{fig:tic-tac-toe}
\vspace{-3mm}
\end{figure}

In order to ensure safety and general correctness, synthesis, either reactive or probabilistic, has emerged as a promising approach to creating correct-by-construction robot strategies \cite{Hadas:ICRA:2007, kress2018synthesis, he2019efficient}. 
In reactive synthesis, the worst-case behavior of the human/environment is considered. This is overly conservative and lacks the power to describe many scenarios where one possibility is known to be more likely than another. Essentially, by assuming the human/environment is \emph{adversarial}, the robot becomes overly pessimistic, resulting in conservative strategies that are ``unfriendly'' and ``competetive'' with the human.
It also drastically limits the scenarios, for which task-completion guarantees can be provided. 
This is the reason previous methods place unwieldy limitations on human~\cite{he2019efficient}, e.g., human takes at most a fixed number of actions.

Probabilistic synthesis has been proposed as an alternative to reactive synthesis to address this issue. Those methods view the human/environment as a \emph{probabilistic} agent \cite{Junges:Arxiv:2016, wells2021probabilistic}.
This can describe many types of human behaviors, not just adversarial behavior; however, it is unrealistic as it assumes that the human behavior is Markovian.
Generally, humans have their own objectives and take actions accordingly. Hence, they should be treated as strategic agents.   For instance, in a tic-tac-toe game, as depicted in \autoref{fig:tic-tac-toe}, both human and robot aim to win.  However, it is not clear if the human is a novice player (makes imperfect moves), a master player (makes perfect moves), or somewhere in between.  Hence, both probabilistic and strategic aspects are present, which a purely probabilistic model cannot capture \cite{abate2021a, wells2021probabilistic}.

To address these limitations, we present robot strategy synthesis using stochastic games. Intuitively, stochastic games can be considered as a generalization of Markov Decision Processes (MDPs), where instead of one agent making decisions, multiple agents make decisions. Stochastic games subsume the expressive power of reactive and probabilistic synthesis; giving us the most general model for robot strategy synthesis. 
For instance, they allow modeling of the scenario in \autoref{fig:tic-tac-toe}, even for a robot with imperfect actuation, e.g., may accidentally drop a piece in an unintended location due to a bad grasp.
The key benefit of stochastic games compared to prior works is not that the models are more accurate (though this can be the case), but rather modeling human-robot manipulation as a stochastic game makes fewer assumptions and is thus more robust.
Because of their expressive power, however, stochastic games bring new challenges in terms of scalability. This paper only begins to address these challenges.



In this work, 
we bridge the gap between robotic manipulation domain and the expressive power of stochastic games.
We mainly focus on the abstraction construction of the continuous manipulation domain in the presence of a human and robot action uncertainty as a discrete two-player stochastic game.  
We present conditions and semantics under which this abstraction can be viewed as a turn-based game, improving computation tractability.  Further, we show that the strong assumption that the human takes a pre-defined number of actions (as in \cite{He:IROS:2017}) can be relaxed in our abstraction.  
We also provide an implementation that enables scalability by bypassing the built-in model construction of stochastic games in the existing tool, namely PRISM-games~\cite{kwiatkowska2020prism}. 
Finally, we illustrate the power of our approach on several case studies and show scalability in a set of benchmarks.


The contributions of this work are fourfold: (i) we formalize how to model the human-robot manipulation domain as turn-based, two-player stochastic game and use existing tools to synthesize optimal strategies for the robot;
(ii) we relax the assumptions on human interventions while still treating the human as a strategic agent; (iii) we improve the scalability of the existing tool and 
provide an open-source tool for efficient synthesis for robotic manipulation scenarios (available on Github~\cite{wells2021github});
(iv) we illustrate the efficacy of our proposed approach on several case studies and benchmarks.




\textbf{Related Work. }
\emph{Synthesis} is the problem of automatically generating a correct-by-construction plan or strategy from a high-level description (specification) of a task. 
The specifications are usually expressed in Linear Temporal Logic (\ltl) \cite{pnueli1977temporal}, and for robotic systems, \ltl interpreted over finite traces (\ltlf) \cite{de2013linear, zhu2017symbolic} is popular due to its ability to describe tasks that need to be completed in finite time. 
When an agent interacts with the world, we are interested in synthesizing a strategy that \emph{reacts} to the environment. Reactive synthesis has been examined as a stand-alone problem as well as in robotics \cite{Hadas:ICRA:2007, kress2018synthesis}. Most works on reactive synthesis for robotics focus on mobile robots \cite{Hadas:ICRA:2007, wongpiromsarn2012receding, vasile2014reactive, wolff2013efficient}, which has a relatively simple state space compared to manipulation.
Reactive synthesis has also been examined for manipulation \cite{He:IROS:2017, He:RAL:2018, Muvvala:ICRA:2022} domains. We build on these later works in this paper. 

Probabilistic synthesis has been examined for general domains \cite{kwiatowska2011prism, baier2008principles, miao2018hybrid, lu2015uav_sg, wells2019ltlf}, including robotic manipulation \cite{wells2021probabilistic}. In the context of learning, stochastic games with unknown transitions have been studied for abstracted robotic systems \cite{bozkurt2020modelfree}. In synthesis, we assume the transitions between states are known a priori. 
Existing works on stochastic synthesis for manipulation domain use MDPs, which only allow one strategic agent. Thus, they assume the human behaves in a mechanical fashion and synthesize an optimal robot policy. Using stochastic games allows us to reason about a strategic human agent.
We focus on modeling human-robot manipulation scenarios with stochastic games where tasks are defined using formal language, which has not been studied.










%% file: sections/problem_setup.tex
\section{Problem Setup}
\label{sec:problem_setup}

\input{tikzfigures/ex_abstraction}

In this work, we focus on a robotic manipulator with ``trembling hands" operating in the presence of a human.  
Given a high-level task for the robot and knowledge on the behaviors of general humans, our aim is to synthesize a strategy for the robot to maximize the probability of completing the task.


\subsection{Probabilistic Abstraction of Manipulation Domain}
We model the manipulation domain as an MDP by abstracting configuration space $C = C_r \times C_o$, where $C_r$ and $C_o$ are the robot and movable objects configuration spaces, respectively. 
Intuitively, a state of this MDP captures relevant features of $C$. That is, the state is a tuple of objects and their locations.
Further, using Planning Domain Definition Language (PDDL) \cite{haslum2019introduction}, we ground and define robot actions along with preconditions and effects of these actions from every state \cite{Muvvala:ICRA:2022}. Since our actions have stochastic outcomes, we define a probability distribution associated with the effects of robot actions. Formally, 

\begin{definition}[Probabilistic Manipulation Domain Abstraction] A probabilistic manipulation domain is an MDP tuple $\mathcal{M} = (S, A, P, s_0, AP, L)$ where,
\begin{itemize}
    \item $S$ is a finite set of states,
    \item $s_0 \in S$ is the initial state,
    \item $A$ is a finite set of robot actions,
    \item $P: S \times A \times S \to [0, 1]$ is the probability distribution over the effects of the robot's action $a \in A$ and $\sum_{s' \in S} P(s, a, s') = 1$ for all state-action pairs,
    \item $AP$ is the set of task-related propositions that can either be true or false, and
    \item $L: S \to 2^{AP}$ is the labeling function that maps each state to a set of $AP$ that are true in $s \in S$.
\label{def:manip_domain_abstraction}
\end{itemize}    
\end{definition}

An execution of MDP $\mathcal{M}$ is a \textit{path}  $\omega = \omega_0 \xrightarrow{a_0} \omega_1 \xrightarrow{a_1} \ldots \xrightarrow{a_{n-1}} \omega_n$, where $\omega_i \in S$, $a_i \in A$, $\omega_0 = s_0$, and $P(\omega_i, a_{i}, \omega_{i+1}) > 0$ for all $0\leq i \leq n-1$.  The set of finite paths is denoted by $S^*$.
The \textit{observation trace} of $\omega$ is $\rho = L(\omega_0) \ldots L(\omega_n)$ the sequence of sets of atomic propositions observed along the way.  
We define the task of the robot according to these observations, below. 
A \textit{robot strategy} $\pi: S^* \to A$ is a function that chooses an $a \in A$ for the robot given the path $\omega \in S^*$ executed so far.


\begin{example}[MDP]
    Consider the continuous manipulation domain in \autoref{fig:toy_example}. The corresponding MDP is depicted in \autoref{fig:example_mdp_robot}. The robot is tasked with building an arch with blue block (not shown) on top. The initial state is defined as $s_0 := \{O_{02}, O_{13}, O_{21}\}$ where $O_{ij}$ corresponds to object $i$ placed at location $j$. Here $O_0, O_1, O_2$ are the red, yellow, and blue blocks, respectively. From the initial state, under the robot-grasp blue block action, there is a 10\% chance of failure and a 90\% chance of success. The alternate action is to grasp the yellow block and finally place blue on top.
    \input{tikzfigures/ex_mdp_only_robot}
\end{example}



\subsection{Manipulation Tasks as \ltlf formulas}

As robotic tasks must be accomplished in finite time, Linear Temporal Logic over finite traces (\ltlf)~\cite{de2013linear} is an appropriate choice for the specification language.  That is because \ltlf is very  expressive (same syntax as \ltl) but its interpretations are over finite behaviors. 

\begin{definition}[\ltlf Syntax]
	Given a set of atomic propositions $AP$, an \ltlf formula is defined recursively as
	\begin{equation*}
		\ltlfphi := \top   \mid   p   \mid   \neg \ltlfphi   \mid   \ltlfphi_1 \wedge \ltlfphi_2   \mid   X \ltlfphi  \mid  \ltlfphi \, U \ltlfphi
	\end{equation*}
	where $p \in AP$ is an atomic proposition, $\top$ (``true''), $\neg$ (``negation'') and $\wedge$ (``conjuction'') are the Boolean operators, and  $X$ (``next'') and $U$ (``until'') are the temporal operators.
\end{definition}

\noindent
The common temporal operators ``eventually'' ($F$) and ``globally'' ($G$) are defined as: $F \, \ltlfphi = \top \, U \, \ltlfphi$ and $G \, \ltlfphi = \neg F\, \neg \ltlfphi$.

The semantics of an \ltlf formula are defined over finite traces in $(2^{AP})^*$ \cite{de2015synthesis}. 
We say a path $\omega \in S^*$ accomplishes $\ltlfphi$, denoted by $\omega \models \ltlfphi$, if its observation trace satisfies $\ltlfphi$.    

\begin{example}[Example \ltlf specification]

The \ltlf formula for constructing an arch from \autoref{fig:toy_example} with any two blocks as support and blue block on top can be written as, 
\begin{align*}
    \phi_\text{arch} = & F \left( p_{\text{block, support}_{1}} \land p_{\text{block, support}_{2}} \land p_\text{blue, top} \right) \land \\
    & \quad G \left( \neg(p_{\text{block, support}_{1}} \wedge p_{\text{block, support}_{2}}) \rightarrow \neg p_\text{blue, top} \right) 
\end{align*}
where $support_i \in \{L_2, L_3\} \; \forall i \in \{1, 2\}$ and $top := L_4 $.
\label{ex:task_phi}
\end{example}

\subsection{Problem Statement}

In this work, we are interested in synthesizing strategies for the robot operating in the presence of human. In our setting, human behavior can be abstracted as human moving objects and the actions can be formalized as object(s) moving from one location to another. 
We aim to develop a general framework; hence, we do not assume knowledge about the particular human, with whom the robot is interacting.

Further, we assume that humans are strategic agents who choose actions according to some objective, which is latent to the robot. Also, we assume general knowledge on the likelihood of the human moving a specific block to some location. 
Such general likelihoods can be inferred from past experiences (data) on various humans.
Our goal is to synthesize a strategy for the robot to maximize the likelihood of achieving its task.


\begin{problem}
\label{problem}
Given a robotic manipulator with its MDP abstraction and \ltlf task formula $\phi$ in the presence of a human with a latent objective and general likelihood of (taking) actions,
\begin{enumerate}
    \item Abstraction: generate a finite abstraction of the interaction between the robot and human through object manipulation such that it captures the strategic and stochastic aspects of both agents,
    \item Synthesis: synthesize a strategy for the robot that maximizes its probability of accomplishing $\phi$.
\end{enumerate}
\end{problem}

There are several challenges in Problem~\ref{problem}. We want to model a strategic human with imperfect decision making capabilities. Also, we want to allow the robot to be a strategic agent that could fail sometimes in executing its action. Hence, the focus of our approach is the construction of an abstraction that captures all the necessary aspects of the problem for best decision making. Additionally, note that both the manipulation domain and reactive synthesis are notorious for their \emph{state-explosion} problem \cite{Muvvala:ICRA:2022, wells2021probabilistic}. Problem~\ref{problem} combines the two; hence, computational tractability is an aspect that we want to ensure in our approach.

%% file: tikzfigures/ex_abstraction.tex
\begin{figure}[t!]
    \centering
    \begin{subfigure}{0.25\columnwidth}
        \includegraphics[width=\textwidth]{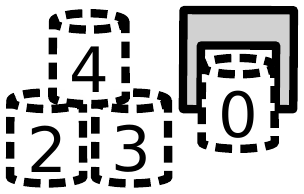}%
        \label{fig:toy_example_setup}
    \end{subfigure}
    \qquad \quad
    \begin{subfigure}{0.25\columnwidth}
        \includegraphics[width=\textwidth]{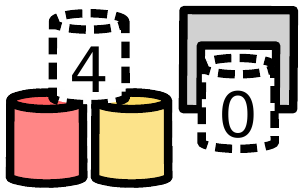}
        \label{fig:toy_example_initial}
    \end{subfigure}
    \caption{Manipulation domain: (left) the locations of interest, where the Else location ($L_1$) contains all objects not otherwise shown. (right) Initial state with red and yellow blocks at $L_2$ and $L_3$ and the blue block at $L_1$.}
    \label{fig:toy_example}
\vspace{-3 mm}
\end{figure}

%% file: tikzfigures/ex_mdp_only_robot.tex
\begin{figure}[t]
    \vspace{0em}
        \centering
        \scalebox{1}{
        \begin{tikzpicture}[auto,node distance=2.5cm,on grid,line width=0.4mm,
                             initial/.style ={initial by arrow, initial left}, initial text=$ $]
            \tikzstyle{round}=[thick,draw=black,circle]
    
            \node[state,initial, inner sep=1pt,minimum size=0pt] (s0) {\includegraphics[width=11mm]{figs/arch_init}};
            \node[state,inner sep=1pt,minimum size=0pt] (s1) [right=of s0] {\includegraphics[width=11mm]{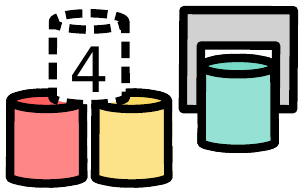}};
            \node[state,inner sep=1pt,minimum size=0pt] (s2) [below=1cm of $(s0)!0.5!(s1)$] {\includegraphics[width=11mm]{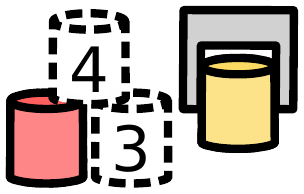}};
            \node[state,accepting,inner sep=1pt,minimum size=0pt] (s3) [right=of s1] {\includegraphics[width=11mm]{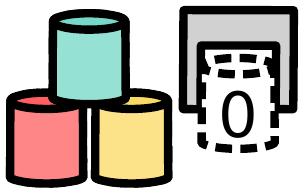}};		
    
            \path[->] 	(s0) edge node [below] {\footnotesize 0.9} (s1)
                        (s0) edge[pos = 0.9] node [below, left] {\footnotesize 0.9} (s2)
                        (s0) edge[out=2, in=12, loop] node [above] {\footnotesize 0.1} (s1)
                        (s0) edge[out=305, in=305+12, loop] node [above right] {\footnotesize 0.1} (s1);
            \path[->] 	(s1) edge [bend right=45, pos=0.8] node [above] {\footnotesize 0.9} (s0)
                        (s1) edge node [below] {\footnotesize 0.9} (s3)
                        (s1) edge[out=2, in=12, loop] node [above]{\footnotesize 0.1} (s1);
            \path[->] 	(s2) edge [bend left=45, pos=0.8] node [below left] {\footnotesize 0.9} (s0)
                        (s2) edge [out=169, in=181, loop] node [below] {\footnotesize 0.1} (s2);
                        
        \end{tikzpicture}
        }
        \caption{Example abstraction of manipulation domain from \autoref{fig:toy_example} with stochasticity for robot actions.
        }
        \label{fig:example_mdp_robot}
        \vspace{-3mm}
\end{figure}

%% file: sections/modeling.tex
\section{Stochastic Game Abstraction}
\label{sec:abstraction}


In this section, we discuss how strategic and stochastic elements of the human and robot are combined to form a two-player stochastic game. Specifically, we deal with a fully observable two-player game.  Naturally, this game is concurrent in the continuous domain, but concurrent games are known to suffer from computational tractability \cite{kwiatkowska2021automatic}.
Our goal is to define semantics that allow a turn-based modeling for the purpose of strategy synthesis such that the execution of the strategy is seemingly concurrent at the runtime. We first formally define a two-player, turn-based stochastic game (simply, \emph{stochastic game}) \cite{condon1992complexity}, and then show our abstraction to this game.



\begin{definition}[Stochastic Game]
A stochastic game is a tuple $G = (S, s_0, A_s, A_e, T, C, AP, L)$, where $S, s_0, AP$ and $L$ are as in \autoref{def:manip_domain_abstraction}, and
\begin{itemize}
    \item $A_s$ and $A_e$ are the finite set of robot \& human actions,
    \item $T: S \times (A_s \cup A_e) \times S \to [0,1]$ is the probabilistic transition relation, and
    \item $C: S \mapsto \{s,e\}$ designates which player controls the choice of action at each state.
\end{itemize}
\label{def:stochastic_game}
\end{definition}
Here, players $s$ (system) and $e$ (environment) are the robot and human, respectively.
An execution of the game $G$ is a sequence of visited states as players take turns in making moves.
The choice of action for Player $i \in \{s,e\}$ is determined by the strategy $\pi_i:S^* \to A_i$ that picks actions according to the execution of the game so far.

For the strategic players, we follow the models of prior work \cite{he2019efficient}. The robot player's actions follow a standard pick-place domain, which typically
can be encoded in PDDL as described above. The human player has the same abilities but is assumed to move relatively quickly compared to the robot.
Additionally, unlike previous approaches, we do allow the human to hold onto an object.
Thus, we model the robot's gripper and the human gripper and make a fairness assumption that the human will eventually return the object.

\textbf{Game States. }
As in \cite{he2019efficient}, we model the continuous world by grouping locations into ``regions of interest''. These include a ``end-effector'' region representing the robot's gripper and an ``Else'' region representing all locations not particularly specified.
To allow the robot to react at any point, the model should be constructed such that every valid arrangement of objects in the real world has game states for both human and robot turns.  These states are equivalent to the robot MDP states in \autoref{def:manip_domain_abstraction}. 

\textbf{Game Actions. }
\label{sec:allocating_turns}
In prior works~\cite{He:IROS:2017,Muvvala:ICRA:2022,Muvvala:IROS:2023}, the human is typically assumed to move faster than the robot, leading to multiple human moves per robot move. We follow this assumption and examine several models of turn allocation. We have several modeling choices and present results for all of them.
The set of robot actions $A_s = A$ is the same set of actions in \autoref{def:manip_domain_abstraction}.  The human actions $A_e$ are every possible move of the objects to the locations of interest, ``Else'', and human's gripper.  Then, the transition relation $T(s,a,s') = P(s,a,s')$ if $a \in A_s$; otherwise, it is obtained from the likelihood of the human actions as discussed below.

In reactive synthesis~\cite{He:IROS:2017}, a limit \emph{k} is placed on the total number of human interventions to ensure the specification is realizable. This limit is unintuitive and somewhat unrealistic. For example, suppose the model assumes the human intervenes at most 30 times ($k = 30$). Then, during execution, once the robot observes the 30th action, it will act as though the human will no longer interfere. Unfortunately, unless there is some external reason for this limit, the robot should arguably assume the human is \emph{more} likely to interfere because it has observed this happen $30$ times already.

We generalize this as a \emph{ratio} of human and robot actions. For example, we could allow 1 human action for every 2 robot actions (denoted by $2:1$). We implement this using counters that reset every time the game changes control from one player to another. So a player cannot ``skip'' turns now in order to take more consecutive turns later. 
While this could be encoded as a two-player game, the ability to express the stochasticity in robot's success rate (of executing actions) or/and the human's tendency to intervene at particular locations can not modeled using a purely game theoretic approach.

Our other model uses a probability of handing control from one player to another. This implies a probabilistic limit but avoids setting a hard limit on action for either player. This is achieved by including an action in $A_e$ that evolves to a state after which the human does not intervene. 
This is a natural weakening of the hard limit on human intervention. Note that this cannot be modeled using MDP where the effects of human actions are inherently random in nature. Thus, stochastic games allows us to reason over strategic players while also considering stochasticity in their execution.

In addition to assigning control of game states to each player, we need some way to turn the continuous, real-time interaction into a turn-based game. We do this as in \cite{He:IROS:2017}, by assuming certain robot actions are ``atomic'' while giving the human actions priority over all non-atomic actions. Here, as in previous papers, the atomic robot actions are grasp and place (not including the transit or transfer preceding the opening / closing of the gripper). Once we have chosen a way to model actions, and under our assumptions about discrete states and atomic action executions, we can create a turn-based stochastic game.

\input{tikzfigures/ex_sg}

\begin{example}
A partial two-player stochastic game for our manipulation domain is shown in \autoref{fig:stochastic_game}. The actions taken in circular states are controlled by the system and those taken in rectangular states are controlled by the environment. The same action could stochastically lead to different possible states. For example, the robot's action from initial state is to 
grasp a block from the initial state and stochastically determine whether to grasp the yellow or the blue block.    
\end{example}



\begin{figure*}[ht]
    \centering
    \begin{subfigure}[t]{0.33\textwidth}
        \centering
        \includegraphics[width=0.9\linewidth]{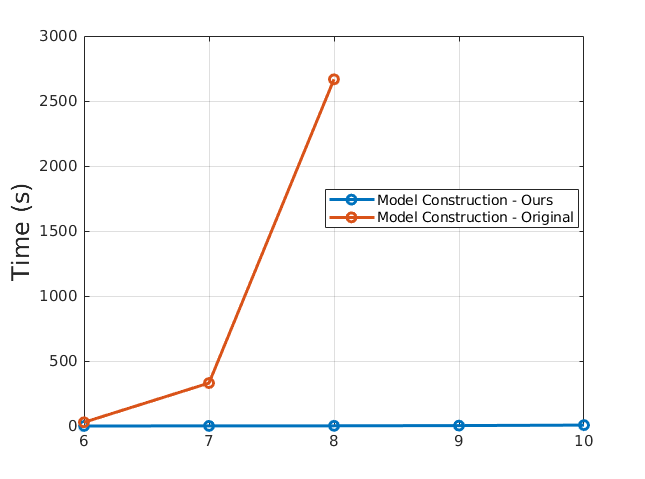}
        \vspace{-2mm}
        \caption{$|O| = 3$, varying $|L|$}
    \label{fig:prob_vary_loc_prism_vs_ours}
    \end{subfigure}%
    ~
    \begin{subfigure}[t]{0.33\textwidth}
        \centering
         \includegraphics[width=0.9\linewidth]{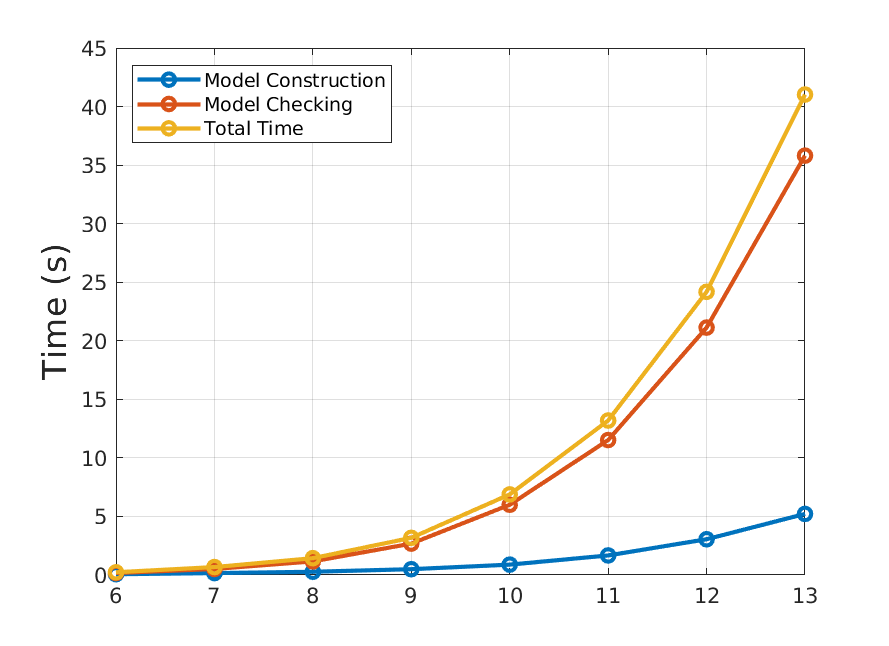}
         \vspace{-2mm}
        \caption{$|O| = 3$,  varying $|L|$}
    \label{fig:unlimited_3_obj_vary_loc}
    \end{subfigure}%
    ~
    \begin{subfigure}[t]{0.33\textwidth}
        \centering
        \includegraphics[width=0.9\linewidth]{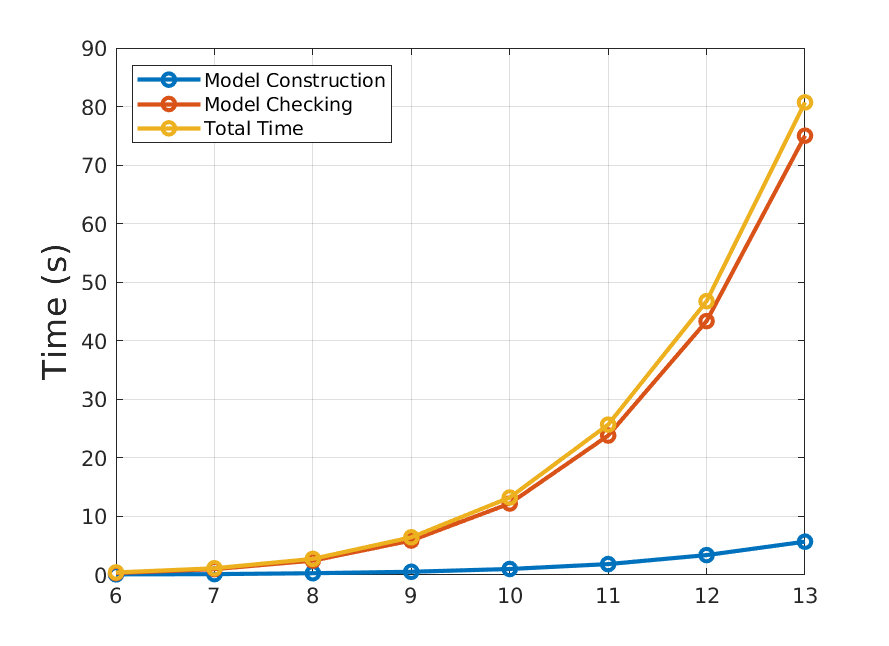}
        \vspace{-2mm}
        \caption{$|O| = 3$, varying $|L|$}
    \label{fig:prob_3_obj_vary_loc}
    \end{subfigure}%
    \\
    \begin{subfigure}[t]{0.33\textwidth}
        \centering
         \includegraphics[width=0.9\linewidth]{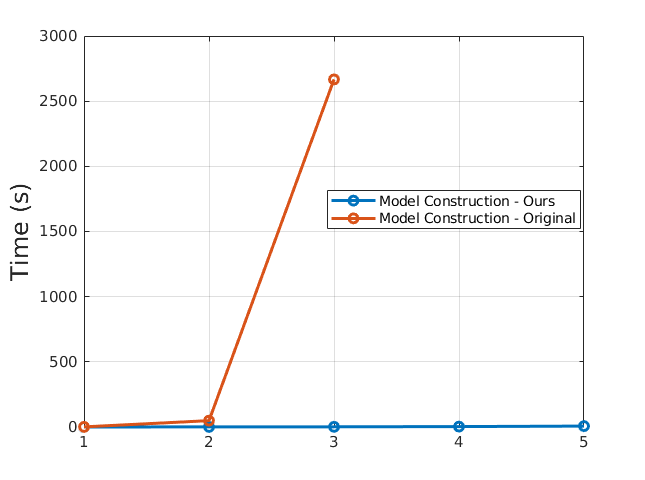}
         \vspace{-2mm}
        \caption{$|L| = 8$, varying $|O|$}
    \label{fig:prob_vary_obj_prism_vs_ours}
    \end{subfigure}%
    ~
    \begin{subfigure}[t]{0.33\textwidth}
        \centering
         \includegraphics[width=0.9\linewidth]{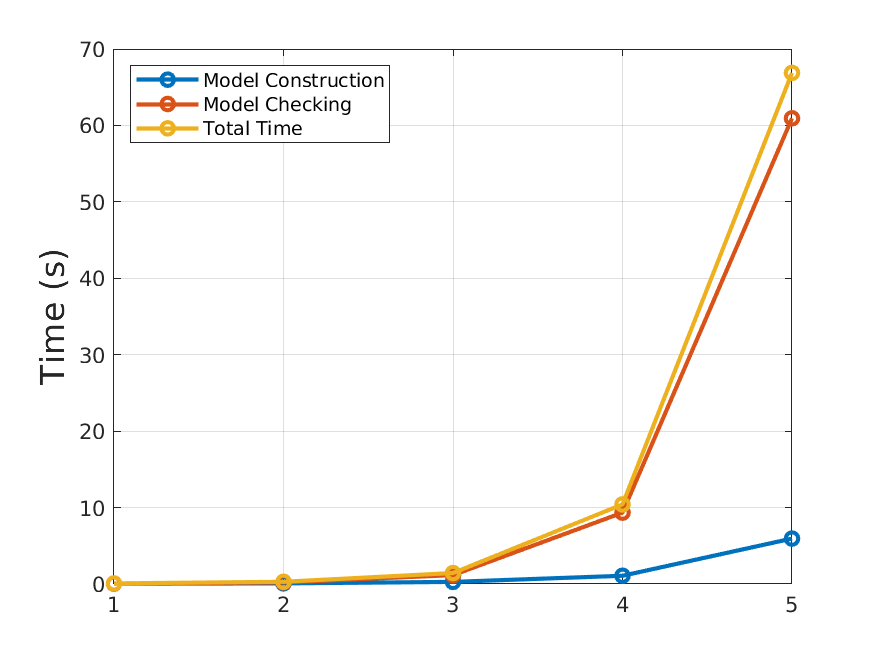}
         \vspace{-2mm}
        \caption{$|L| = 8$, varying  $|O|$}
    \label{fig:unlimited_8_locj_vary_obj}
    \end{subfigure}%
    ~
    \begin{subfigure}[t]{0.33\textwidth}
        \centering
         \includegraphics[width=0.9\linewidth]{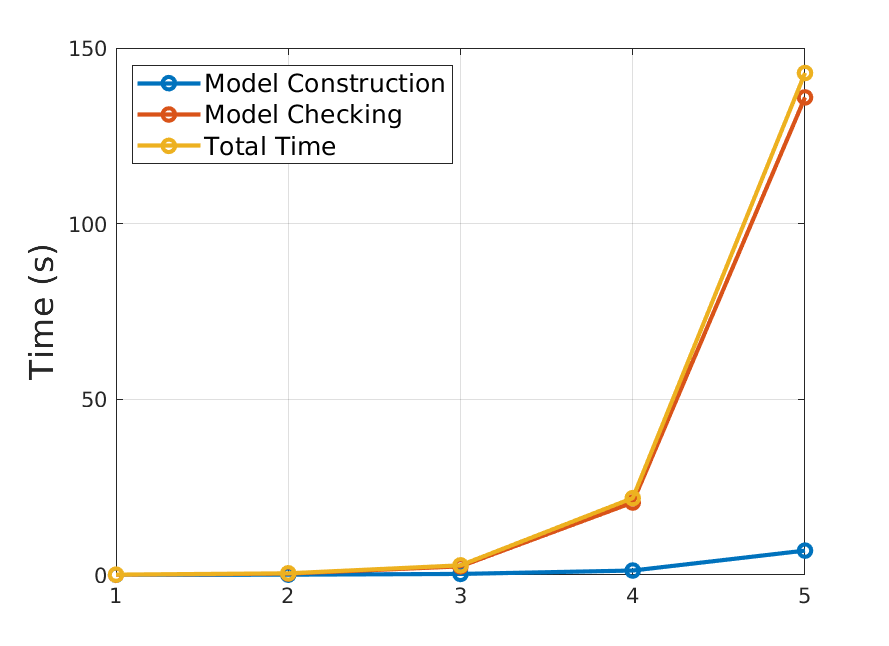}
         \vspace{-2mm}
        \caption{$|L| = 8$, varying $|O|$}
    \label{fig:prob_8_locj_vary_obj}
    \end{subfigure}%
\caption{Benchmark results for different scenarios using our approach. (a) and (d) illustrate the model construction time using the original PRISM-games and our implementation for the probabilistic human termination scenario. (b) and (e) illustrate computation times for the 1:1 action ratio scenario, and (c) and (f) correspond to the probabilistic human termination scenario.}
\label{fig: benchmark_figs}
\end{figure*}

\begin{remark}
    Winning the game translates to finishing task $\phi$, and winning strategies are strategies that guarantee task completion. Termination of the game is typically defined as reaching an accepting or violating finite prefix of a trace. This could be insufficient in certain cases, e.g., a robot asked to tidy a room will stop once the room is cleaned even if the human is approaching some object to displace it. Our game modeling allows the human and robot to ``negotiate'' termination so that the robot only considers the task complete when the human agrees.
\end{remark}


%% file: tikzfigures/ex_sg.tex
\begin{figure}[t!]
        \centering
        \begin{tikzpicture}[auto,node distance=3cm,on grid,line width=0.4mm,
                            initial/.style ={initial by arrow, initial above},
                            initial text=$ $]
          \tikzstyle{round}=[thick,draw=black,circle]
    
            \node[state, initial, inner sep=1pt,minimum size=0pt] (s0) {\includegraphics[width=12mm]{figs/arch_init}};
            \node[draw=none, minimum size=0pt] (s1) [right=2cm of s0] {}; 
            \node[rectangle,draw,inner sep=4pt,minimum size=0pt] (s8) [above=1.5 cm of s1] {\includegraphics[width=13mm]{figs/arch_other}};
            \node[rectangle,draw,inner sep=4pt,minimum size=0pt] (s2) [right=1.5cm of s1] {\includegraphics[width=13mm]{figs/arch_grasped}};
            \node[draw=none, rectangle, minimum size=0pt] (s3) [above right=2 cm of s2] {};
            \node[draw=none, rectangle, minimum size=0pt] (s4) [right=2 cm of s2] {};
            \node[state,draw,inner sep=4pt,minimum size=0pt] (s5) [right=1.5cm of s4] {\includegraphics[width=10mm]{figs/arch_grasped}};
            \node[state,draw,inner sep=4pt,minimum size=0pt] (s6) [above right=2cm of s3] {\includegraphics[width=10mm]{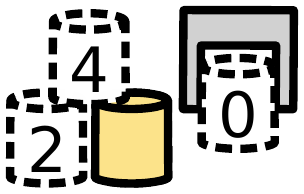}};
            \node[state,draw,inner sep=4pt,minimum size=0pt] (s7) [above left=2cm of s3] {\includegraphics[width=10mm]{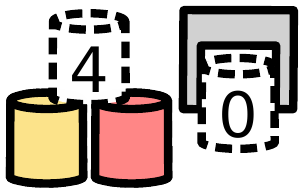}};
            \node[draw=none,fill=none, right=1.5cm of s7] {\Large\dots};\
            \node[] (s9) [left=2 cm of s7] {};
             \node[] (s10) [left=1.5 cm of s9] {};
             \node[draw=none,fill=none, right=-0.5cm of s9] {\Large\dots};\

            \path[->]
                (s0) edge [bend right] (s8)
                (s0) edge [in=180,out=0] (s2)
                (s2) edge [bend right, out=0] (s7)
                (s2) edge [bend left, out=0, in=230] (s6)
                (s2) edge (s5)
                (s8) edge [bend right] (s9)
                (s8) edge [bend left] (s10);
            
        \end{tikzpicture}
        \caption{Stochastic game variant of MDP in \autoref{fig:example_mdp_robot}. The circle and rectangle states belong to the robot and human player. For this e.g. we allow human to move objects from the robot's gripper. The top row shows multiple human movements, while the state on the right corresponds to no human intervention.}
        \label{fig:stochastic_game}
    \vspace{-5mm}
\end{figure}

%% file: sections/synthesis.tex
\section{Strategy Synthesis}


For a given \ltlf\ specification, synthesis reduces to solving a stochastic game for a reachability objective, i.e., reach a target state \cite{kwiatkowska2020prism}.
That game is the composition of $G$ with the automaton that is constructed from $\phi$ \cite{de2013linear}. Solving stochastic games with reachability objectives lies in the complexity class NP $\cap$ coNP\cite{condon1993algorithms}.
PRISM-games makes use of the model checking algorithms described in \cite{kwiatkowska2020prism}, that relies on Value Iteration 
to compute the values for all states of the game \cite{Chatterjee2008VI}. The algorithm can be decoupled into two stages.


\paragraph*{Precomputation Stage}
During this stage, we identify states of the game for which the probability of satisfaction is 0 or 1, and the maximal end components of the game. Informally, an end component is a set of states for which, there exists a robot strategy such that it is possible to remain forever in that set once entered. Efficiency and accuracy can be improved by using this step.
Next, numerical computation is performed on the remaining states in the game.

 \paragraph*{Numerical computation stage}
The probability of reaching a target state is $1$ if the state belongs to the target end component, else we iteratively update state values until we reach a fixed point. At every iteration, we perform  $\max_a (\sum_{s’} T(s,a, s') \cdot p(s’))$ if $s$ belongs to robot player else we perform $\min_a(\sum_{s’} T(s,a, s') \cdot p(s’))$.
Here $T$ is from \autoref{def:stochastic_game}, and $s, a, s'$ are the current state, action, and successor state, respectively. $p(s')$ denotes the value associated with the successor state in the previous iteration. 
 While PRISM-games is a mature toolbox, the implementation for solving stochastic games is less mature than tooling for MDPs, and we found the bottleneck to be model construction rather than Value Iteration. Below, we discuss how we mitigate this bottleneck.



%% file: sections/results.tex
\section{Implementation and Results}
Here, we present benchmarks based on experiments from \cite{wells2021probabilistic}. We run our tests using PRISM-games \cite{kwiatkowska2020prism} and discuss our modifications to remove a performance bottleneck when importing models.
All the experiments are run on an Intel i5 -13th Gen 3.5 GHz CPU with 32 GB RAM. The tool is available on GitHub \cite{wells2021github}. The results are shown in \autoref{fig: benchmark_figs}.

\begin{figure*}[t!]
  \centering
  \includegraphics[width=.97\textwidth]{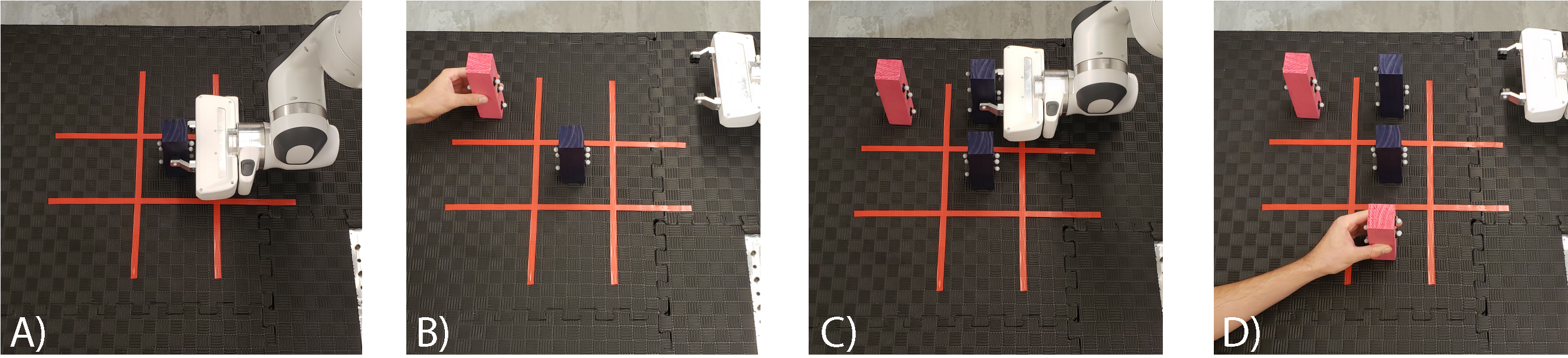}
  \caption{The game begins with (A) and (B). In state (C) the robot chooses a move that maximizes the probability of human failure under the ``trembling hand'' model. In (D) the human will likely place the object in the bottom center, but has two open neighbor cells. Under the robot strategy, the human will have three more chances to fail (video: \small{\url{https://youtu.be/UUBW7QEw6Ng}}).}
  \label{fig:tic_tac_toe_trace}
\vspace{-1mm}
\end{figure*}

\textbf{Scalability. }
We test a simple pick-and-place manipulation domain, varying the number of objects and locations. Three locations
are reserved for the robot and human gripper, and the terminal location for each player. Only one object can be manipulated by the robot and human, while multiple objects can be placed at other locations. The task is to place objects in their desired locations. 

PRISM's default configuration reads in modeling files written in the PRISM modeling language. As PRISM is (primarily) a symbolic engine, a structured, hierarchical model is preferred as it exploits regularity in the abstraction. Our model is naturally flat, and hence PRISM modeling language suffers from scalability.
Therefore, we implement functionality to import the stochastic games models through the direct specification of their transition matrix, state, label, and player vectors.

We use a Python script to automate the construction of the model files for direct specification of transition matrix and state vectors outside PRISM and then import them in PRISM-games.
We benchmark this method of model construction as shown in \autoref{fig:prob_vary_loc_prism_vs_ours} and \autoref{fig:prob_vary_obj_prism_vs_ours}. We see that while the PRISM's original implementation (in red) fails to scale beyond 3 objects and 8 locations, our approach (in blue) not only scales beyond this bottleneck but is also 2 orders of magnitude faster.

We also present benchmarks based on the modeling choices described in \autoref{sec:abstraction}.
\autoref{fig:unlimited_3_obj_vary_loc} and \autoref{fig:unlimited_8_locj_vary_obj} correspond to model construction and synthesis time for 1:1 scenario where we allow one human action for every robot action. In this scenario, the human could potentially undo every action the robot does and hence the robot cannot guarantee task completion. We see that the computation time grows exponentially as the state space increases for fixed $|O|$ and varying $|L|$ and vice versa.

\begin{table}[]
\caption{Abstraction \& Synthesis comp. times for 3 objects.}
\centering
\resizebox{1\columnwidth}{!}{%
\begin{tabular}{c||c|c|c|c|c}
\toprule
Case & $|L|$ & States & Transitions & Model Const.(s) & Synthesis (s)\\ 
 Study & & & & & \\ \hline \hline
\multirow{4}{*}{1:1} & 7 & 8,480 & 42,240 & 0.161 & 0.515 \\
 & 9 & 43,848 & 298,080 & 0.496 & 2.679 \\
 & 11 & 148,608 & 1,288,704 & 1.671 & 11.522 \\
 & 13 & 393,800 & 4,166,400 & 5.216 & 21.126 \\ \hline \hline
\multirow{4}{*}{Prob} & 7 & 9200 & 63440 & 0.148 & 1.002 \\
 & 9 & 45,864 & 442,008 & 0.551 & 5.885 \\
 & 11 & 152,928 & 1,905,120 & 1.862 & 23.822 \\
 & 13 & 401,720 & 6,156,920 & 5.692 & 75.021 \\
\bottomrule
\end{tabular}
\vspace{-5mm}
}
\label{table:abstraction_results}
\end{table}

\begin{table}[]
\caption{Abstraction \& Synthesis comp. times for 5 objects.}
\centering
\resizebox{1\columnwidth}{!}{%
\begin{tabular}{c|c|c|c|c}
\toprule
$|L|$ & States & Transitions & Model Const. (s) & Synthesis (s)\\ \hline \hline
 4 & 148 & 387  & 0.052 & 0.037 \\
 5 & 4,896 & 17,184 & 0.248 & 0.688 \\
 6 & 43,416 & 190,107 & 1.101 & 7.366 \\
 7 & 217,600 & 1,144,320 & 4.406 & 38.768 \\ 
 8 & 787,500 & 4,846,875 & 5.944 & 60.933 \\ 
 9 & 2,304,288 & 16,280,352 & 90.314 & 677.584 \\ 
\bottomrule
\end{tabular}
\vspace{-7mm}
}
\label{table:abstraction_results_2}
\end{table}

We also benchmark scenarios where there is 5\% chance of human termination 
at every state.  In contrast to \cite{He:IROS:2017,he2019efficient}, where a parameter $k$ was used to constrain the number of human interventions, this approach allows greater flexibility and a more intuitive model while still allowing the robot to win the game. The computation times are shown in \autoref{fig:prob_3_obj_vary_loc} and \autoref{fig:prob_8_locj_vary_obj}. Similar to the previous scenario, computation times grow exponentially as the state space increases. For both scenarios, the model construction time, while increasing, is relatively small compared to the synthesis time.

For all of the experiments using our modified implementation, PRISM-games required at most 8 GB of RAM. \autoref{table:abstraction_results} reports the size of the game and time for model construction and strategy synthesis. \autoref{table:abstraction_results_2} illustrates how the abstraction grows for the 1:1 scenario.
The Python script runs out of memory for 5 objects and 10 locations for both scenarios. 

\textbf{Physical Experiment: Tic-Tac-Toe with ``Trembing Hand". }
 Recall the game of tic-tac-toe in \autoref{fig:tic-tac-toe}, where human and robot players alternate turns placing markers in empty cells. Tic-tac-toe can be solved using min-max, but only under the assumption that the robot does not fail to complete an action, i.e., reactive synthesis cannot capture stochasticity in the robot's ability to correctly place a marker at its desired location as per the strategy. While an MDP can capture the stochastic outcomes, it cannot model the strategic nature of the human. On the other hand, using the stochastic games model, we can account for the strategic and stochastic nature of both players. \autoref{fig:tic_tac_toe_trace} illustrates a run of the game.\footnote{Video of more runs: \url{https://youtu.be/UUBW7QEw6Ng}}

In our experiment, we have stochasticity in placing the marker for both the robot and the human. We model this probability of marker placements (say a normal distribution with standard deviation of 1-cell width) as the uncertainty in the player's action. We restrict both the robot and human to not be able to place the marker outside the cells or in an already occupied cell.
Stochastic games allows us to reason over possible human and robot failure. Depending on the task, the robot can move so as to either maximize its chances of winning the game or the chances of human failure. In both case studies, the robot starts the game. We specify these tasks as 
$P_{\max=?}[F(\text{``RobotWin"})]$ and $P_{\min=?}[F (\text{``HumanWin"})]$.
The emergent behavior for the robot under specification 1 is to initially place its marker in the middle. This maximizes its chances of winning while reducing the number of unoccupied cells, which reduces the probability of failure in future robot actions. As the game progresses, we observe that the robot places its marker near crowded cells with fewer empty cells around it. 
For the second specification, we observe that the optimal action, initially, for the robot is to place its marker in the middle. Next, the robot places markers to maximize its winning probability while ensuring as many empty locations as possible for the next optimal human action.


%% file: sections/conclusion.tex
\section{Conclusion}
We present a framework for robot manipulation based on stochastic games. Stochastic games subsume the expressivity of reactive and probabilistic synthesis proposed in previous works.  We illustrate the efficacy of our approach through various scenarios and discuss the emergent behavior. 
Future work should examine symbolic approach to scale to more objects and locations.
Additional work can examine modeling of uncertain observations, reasoning over other agents strategies, concurrent games or games with varying rewards.
